\documentclass[10pt,twocolumn,letterpaper]{article}

\usepackage{cvpr}
\usepackage{times}
\usepackage{epsfig}
\usepackage{graphicx}
\usepackage{amsmath}
\usepackage{amssymb}
\usepackage{tablefootnote}

\usepackage{algorithm}%
\usepackage{algpseudocode}%
\usepackage{multirow}
\usepackage{CJK}

\usepackage{subfig}

\usepackage{color}

\usepackage[breaklinks=true,bookmarks=false]{hyperref}
\graphicspath{./figure}

\cvprfinalcopy 

\def\cvprPaperID{23} 

\begin{document}

\title{Combining ConvNets with Hand-Crafted Features for Action Recognition\\ Based on an HMM-SVM Classifier}

\author{Pichao Wang$^{\rm 1}$\thanks{Both authors contributed equally to this work}, Zhaoyang Li$^{\rm 2}$\footnotemark[1], Yonghong Hou$^{\rm 2}$\thanks{Corresponding author} and Wanqing Li$^{\rm 1}$\\
$^{\rm 1}$Advanced Multimedia Research Lab, University of Wollongong, Australia\\
$^{\rm 2}$School of Electronic Information Engineering, Tianjin University, China\\
{\tt\small pw212@uowmail.edu.au,lizhaoyang@tju.edu.cn, houroy@tju.edu.cn, wanqing@uow.edu.au}\\
}

\maketitle

\begin{abstract}

This paper proposes a new framework for RGB-D-based action recognition that takes advantages of hand-designed features from skeleton data and deeply learned features from depth maps, and exploits effectively both the local and global temporal information.  Specifically, depth and skeleton data are firstly augmented for deep learning and making the recognition insensitive to view variance. Secondly, depth sequences are segmented using the hand-crafted features based on skeleton joints motion histogram to exploit the local temporal information. All training se gments are clustered using an Infinite Gaussian Mixture Model (IGMM) through Bayesian estimation and labelled for training Convolutional Neural Networks (ConvNets) on the depth maps. Thus, a depth sequence can be reliably encoded into a sequence of segment labels. Finally, the sequence of labels is fed into a joint Hidden Markov Model and Support Vector Machine (HMM-SVM) classifier to explore the global temporal information for final recognition.

\end{abstract}

\section{Introduction}

Recognition of human actions from RGB-D (Red, Green, Blue and Depth) data has attracted increasing attention in computer vision in recent years due to the advantages of depth information over conventional RGB video, e.g. being insensitive to illumination changes and reliable to estimate body silhouette and skeleton~\cite{Shotton2011}. Since the first work of such a type~\cite{li2010action} reported in 2010, many methods~\cite{wang2012mining,xia2012view,bloom2012g3d,Oreifej2013} have been proposed based on specific hand-crafted feature descriptors extracted from depth and/or skeleton data and many benchmark datasets were created for evaluating algorithms. With the recent development of deep learning, a few methods~\cite{pichao2015,pichaoTHMS,du2015hierarchical} have been developed based on Convolutional Neural Networks (ConvNets) or Recurrent Neural Network (RNN). However, in most cases, either deeply learned or hand-crafted features have been employed.  Little study was reported on the advantages of using both features simultaneously.

This paper presents a novel framework that combines deeply learned features from the depth modality through ConvNets and the hand-crafted features extracted from skeleton modality. The framework overcomes the weakness of ConvNets being sensitive to global translation, rotation and scaling and leverages the strength of skeleton based features, e.g. Histogram of Oriented Displacement (HOD)~\cite{Gowayyed2013_HOD}, being invariant to scale, speed and clutter of background. In particular, depth and skeleton data are firstly augmented for deep learning and making the recognition insensitive to view variance. Secondly, depth sequences are segmented using the hand-crafted features based on joints motion histogram to exploit the local temporal information. All training segments are clustered using an Infinite Gaussian Mixture Model (IGMM) through Bayesian estimation and labelled for training Convolutional Neural Networks (ConvNets) on the depth maps. Thus, a depth sequence can be reliably encoded into a sequence of segment labels. Finally, the sequence of labels is fed into a joint Hidden Markov Model and Support Vector Machine (HMM-SVM) classifier to explore the global temporal information for final recognition. 
The proposed framework has demonstrated a novel way in effectively exploring the spatial-temporal (both local and global) information for action recognition and has a number of advantages compared to conventional discriminative methods in which temporal information is often either ignored or weekly encoded into a descriptor and to generative methods in which temporal information tends to be overemphasized, especially when the training data is not sufficient. Firstly, the use of skeleton data to segment video sequences into segments makes each segment have consistent and similar movement and, to some extent, be semantically meaningful (though this is not the intention of this paper) since skeletons are relatively high-level information extracted from depth maps and each part of the skeleton has semantics. Secondly, the ConvNets trained over DMMs of depth maps provides a reliable sequence of labels by considering both spatial and local temporal information encoded into the DMMs. Thirdly, the use of HMM on the sequences of segment labels explores the global temporal information effectively and the SVM classifier further exploits the discriminative power of the label sequences for final classification. 
     
The reminder of this paper is organized as follows. Section 2 describes the related work. Section 3 presents the proposed framework. 

\section{Related Work}

Human action recognition from RGB-D data has been extensively researched  and much progress has been made since the seminal work~\cite{li2010action}.  One of the main advantages of depth data is that they can effectively capture 3D structural information.  Up to date, many effective hand-crafted features have been proposed based on depth data, such as Action Graph (AG)~\cite{li2010action}, Depth Motion Maps (DMMs)~\cite{Yang2012a}, Histogram of Oriented 4D Normals (HON4D)~\cite{Oreifej2013}, Depth Spatio-Temporal Interest Point (DSTIP)~\cite{xia2013spatio} and Super Normal Vector (SNV)~\cite{yangsuper}. Recent work~\cite{pichao2015} also showed that features from depth maps can also be deeply learned using ConvNets. 

Skeleton data which is usually extracted from depth maps~\cite{Shotton2011} provides a high-level representation of human motion. Many hand-crafted skeleton based features have also been developed in the past. They include EigenJoints~\cite{Yang2012}, Moving Pose~\cite{zanfir2013moving}, Histogram of Oriented Displacement (HOD)~\cite{Gowayyed2013_HOD}, Frequent Local Parts (FLPs)~\cite{pichao2014} and Points in Lie Group (PLP)~\cite{vemulapalli2014human}, which are all designed by hand. Recently, the work~\cite{du2015hierarchical} demonstrated that features from skeleton can also been directly learned by deep learning methods. However, skeleton data can be quite noisy especially when occlusion exists and the subjects are not in standing position facing the RGB-D camera.

Joint use of both depth maps and skeleton data have also been attempted. Wang et al.~\cite{wang2012mining} designed a 3D Local Occupancy Patterns (LOP) feature to describe the local depth appearance at joint locations to capture the information for subject-object interactions. In their subsequent work, Wang et al. \cite{wang2014learning} proposed an Actionlet Ensemble Model (AEM) which combines both the LOP feature and Temporal Pyramid Fourier (TPF) feature. Althloothi et al. \cite{Althloothi2014} presented two sets of features extracted from depth maps and skeletons and they are fused at the kernel level by using Multiple Kernel Learning (MKL) technique. Wu and Shao \cite{wu2014deep} adopted Deep Belief Networks (DBN) and 3D Convolutional Neural Networks (3DCNN) for skeletal and depth data respectively to extract high level spatial-temporal features. 

\section{Proposed Framework}
\begin{figure}[!ht]
\begin{center}
{\includegraphics[height = 100mm, width = 85mm]{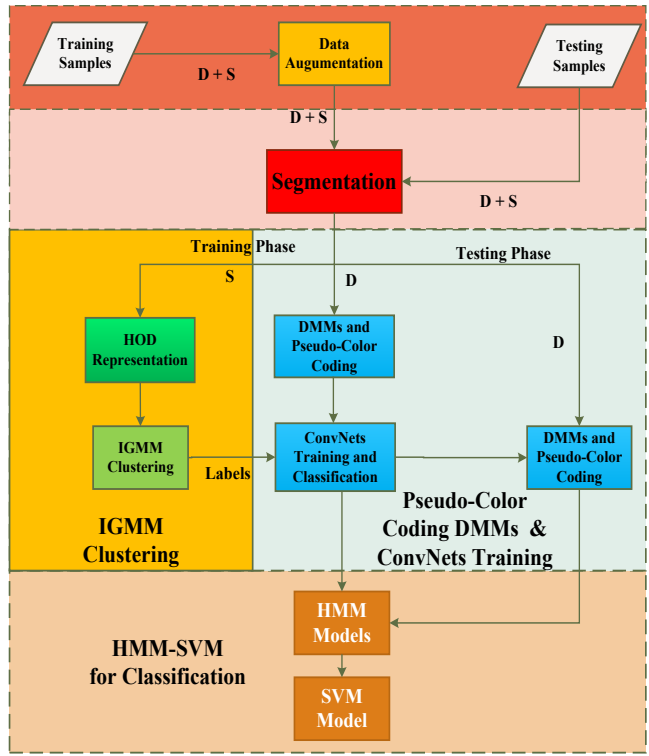}}
\end{center}
\caption{The proposed action coding framework, where D represents Depth data while S denotes Skeleton data.}
\label{fig:framework}
\end{figure}

Fig.~\ref{fig:framework} shows the block-diagram of the proposed framework. It consists of five key components: {\it Data augmentation} to enlarge the training samples by mimicking virtual cameras through rotating the viewpoints; {\it Segmentation} to segment sequences of depth maps into segments by extracting the key-frames from skeleton data, to exploit the local temporal information. {\it IGMM clustering} to label all training segments through clustering; {\it ConvNets on DMMs} to train ConvNets to classify segments reliably; and {\it HMM-SVM} to model the global temporal information of actions and classify a sequence of segment labels into an action.

\subsection{Data Augmentation}
The main purposes of data augmentation are to address the issue of training ConvNets on a small dataset and to deal with view variations. The method presented in~\cite{pichao2015} is adopted in this paper, where depth data is augmented by rotating the 3D cloud points captured in the depth maps and skeleton to mimic virtual cameras from different viewpoints.  More details can be found in~\cite{pichao2015}.


%




\subsection{Segmentation}
In order to exploit the local temporal information, depth and skeleton sequences are divided into segments such that motion across frames within each segment is similar and consistent.  To this end, key-frames are first extracted using the inter-frame and intra-frame joint motion histogram analysis, a method similar to the one described in~\cite{shao2009motion} . The joint motion histograms are insensitive to the background motion compared to the use of optical flow in~\cite{shao2009motion}. Specifically, skeleton data are firstly projected on to three orthogonal Cartesian planes. The motion vectors calculated between two frames of the corresponding joints in each plane are quantized by its magnitude and orientation. The combination of magnitude and orientation corresponds to a bin in the motion histogram. Given the number of joints $J$, the probability of the $k^{th}$ bin in the histogram of one projection is given as:
 

\begin{equation}\label{eq2}
p(k) = \dfrac{h(k)}{J\times3}
\end{equation}

where $h$ denotes the counts of the $k^{th}$ bin. The final motion histogram is a concatenation of the histograms in the three projections. Thus, the entropy of motion vectors in this frame can be defined as:

\begin{equation}\label{eq2}
\eta = -\sum\limits_{k=1}^{k_{max}}p(k)\cdot log_{2}(p(k))
\end{equation}
where $k$ denotes the bin index and $k_{max}$ is the total bin number in the histogram. Intuitively, a peaked motion histogram contains less motion information thus produces a low entropy value; a flat and distributed histogram includes more motion information and therefore yields a high entropy value. 
Then, we follow the work~\cite{shao2009motion} where intra-frame and inter-frame analysis (more details can be found in~\cite{shao2009motion}) are designed to extract key frames such that the extracted key frames contain complex and fast-changing motion and, thus, are salient with respect to their neighboring frames. The details of the algorithm are summarized in Algorithm~\ref{alg1}.   

\begin{algorithm}[htb]
  \caption{ Key Frames Extraction \label{alg1}}
  
  \begin{algorithmic}[1]
    
    \State Select initial frames: $Initial_{i} = \{fx^{1}_{i}, fx^{2}_{i},..., fx^{n_{i}}_{i}\}$ from video $i$ by picking local maxima in the entropy curve 
 calculated by the concatenated motion histogram, where $n_{i}$ denotes the number of initial frames extracted from video $i$; 
 \State Calculate the histogram intersection values between neighboring frames;
 \State Weight the entropy values of $Initial_{i}$ by corresponding histogram intersection values;
 \State Extract key frames $Key_{i} = \{fy^{1}_{i}, fy^{2}_{i},..., fy^{m_{i}}_{i}\}$ by finding peaks in the weighted entropy curve, where $m_{i}$ denotes the number of key frames extracted from video $i$;
  \end{algorithmic}
\end{algorithm}
If $p$ key frames are extracted from each action sample, the whole video sequence can be divided into $M = p + 1$ segments, with the key frames being the beginning or the ending frames of each segment together with the first and last frames of the sequence.

%

\subsection{IGMM Clustering} 

The segments of training samples are clustered using HOD~\cite{Gowayyed2013_HOD} features extracted from the skeleton data and these segments are then labelled and used to train ConvNets on DMMs constructed from the depth maps of segments. Assume that all the action samples in one dataset are segmented to totally $W$ video segments, and let $X = [\vec{x}_{1}, \vec{x}_{2},...,\vec{x}_{W}]$ be the HODs of these segments, where the dimension of HOD is $n$ and $\vec{x}_{l} = [x^{1}_{l}, x^{2}_{l},...,x^{n}_{l}]^{T}$. 

In this paper, it is assumed that the HODs from all segments can be modeled by an IGMM~\cite{wood2006non}. Bayesian approach is adopted to find the best model of $K$ Gaussian components that gives the maximum posterior probability (MAP), each Gaussian accounts for a distinct type or class of segments.  Compared with traditional $K$-means, IGMM dynamically estimates the number of clusters from the data.  Mathematically, the model is specified with the following notations.

\begin{equation}
\begin{array}{c}
c_{i}|\vec{\pi} \sim Multinomial(\cdot|\vec{\pi})\\
\vec{x_{l}}|c_{i} = k, \Theta \sim N(\cdot|\theta_{k})
\end{array}
\end{equation} 

where $C = \{c_{i}\}_{i = 1}^{N}$ indicates which class the HOD belongs to, $\Theta = \{\theta_{k}\}_{k=1}^{K}$, $\theta_{k} = \{\vec{\mu}_{k}, \Sigma_{k}\}$ are distribution parameters of class, and $\vec{\pi} = \{\pi_{k} \}_{k=1}^{K}$, $\pi_{k} = P(c_{i} = k)$ are mixture weights. Here, we do not know the number of clusters $K$, otherwise the complete data likelihood can be computed.
%

To address this problem, a fully Bayesian approach was adopted instead of conventional maximum likelihood (ML) approach, where the relationship between observed data $X$ and a model $H$ in Bayes's rule is:

\begin{equation}
P(X|Y) \propto P(h)P(X|H)
\end{equation}

With respect to the conjugate priors for the model parameters, the same method as the one proposed in~\cite{wood2006non} is adopted, that is  Dirichlet for $\vec{\pi}$ and Normal Times Inverse Wishart for $\theta$~\cite{fraley2007bayesian}~\cite{gelman2014bayesian}.

\begin{equation}
\begin{array}{c}
\vec{\pi}|\alpha \sim Dirichlet(\cdot|\dfrac{\alpha}{K}, \dfrac{\alpha}{K}, ..., \dfrac{\alpha}{K})\\
\Theta \sim G_{0}
\end{array}
\end{equation}

$\Theta \sim G_{0}$ is shorthand for 

\begin{equation}
\begin{array}{c}
\sum_{k} \sim Inverse - Wishart_{v_{0}}(\wedge_{0}^{-1})\\
\vec{\mu}_{k} \sim N(\vec{\mu}_{0}, \sum_{k}/K_{0})
\end{array}
\end{equation}

where $\dfrac{\alpha}{K}$ controls how uniform the class mixture weights will be; the parameters, $\wedge_{0}^{-1}$, $v_{0}$, $\vec{\mu_{0}}$, $K_{0}$ encode the prior experience about the position of the mixture densities and the shape; the hyper-parameters $\wedge_{0}^{-1}$, $v_{0}$ affect the mixture density covariance; $\vec{\mu_{0}}$ specifies the mean of the mixture densities, and $K_{0}$ is the number of pseudo-observations~\cite{gelman2014bayesian}. 

With the model defined above, the posterior distribution can be represented as:

\begin{equation}
\begin{array}{l}
P(C, \Theta, \vec{\pi}, \alpha | X) \propto\\ P(X|C,\Theta)P(\Theta|G_{0})\prod \limits_{i=1}^{N} P(c_{i}|\vec{\pi})P(\vec{\pi}|\alpha)P(\alpha)
\label{eq:igmm-bayesian}
\end{array}
\end{equation}

With some manipulations, Eq.~\ref{eq:igmm-bayesian} can be solved using Gibbs sampling~\cite{neal2000markov} and the Chinese restaurant process~\cite{ghahramani2005infinite}. Details can be found in ~\cite{wood2006non,gelman2014bayesian}.

Through the IGMM clustering, the numbers of active clusters will be estimated and all segments can be labelled with its corresponding cluster through hard assignment. These labelled segments will be the training samples for the ConvNets in the framework.

\subsection{Pseudo-Color Coding of DMMs \& ConvNets Training}
\begin{figure}[!ht]
\begin{center}
{\includegraphics[height = 80mm, width = 85mm]{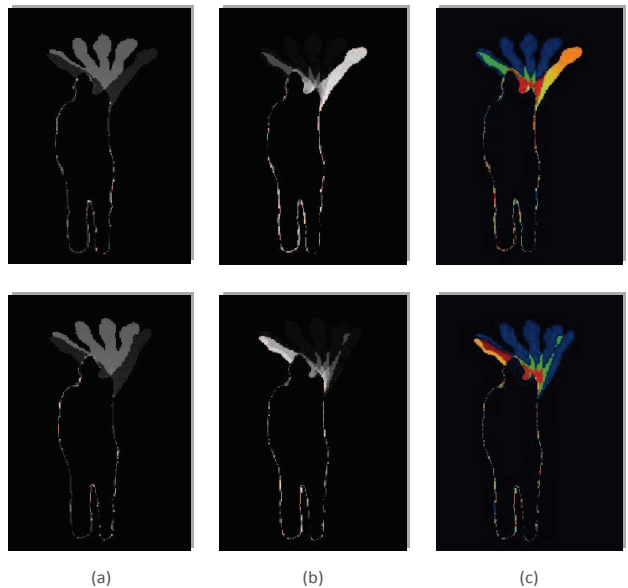}}
\end{center}
\caption{The first row of DMMs represents an action that wave hand from right to left, the second row of DMMs represents an action that wave hand from left to right. (a) are computed in traditional way, (b) are computed in proposed way and (c) are pseudo-coloring coded from (b). }
\label{fig:DMM}
\end{figure}

{\it DMMs and Pseudo-Color Coding} Since skeleton data are often noisy, ConvNets are trained on DMMs~\cite{Yang2012a} of segments for reliable classification from the training segments labelled by the IGMM cluster. Traditional DMMs~\cite{Yang2012a} are computed by adding all the absolute differences between consecutive frames in projected planes, which dilutes the temporal information. After segmentation, it is likely that there are more action pairs such as hands up and hands down within the segments. To distinguish the action pairs, we stack the motion energy within a segment with a special treatment to the first frame as described in Eq.~\ref{dmm}. 

\begin{equation} \label{dmm}
DMM_{v} = map_{v}^{1} + \sum\limits_{k=2}^{M-1}|map_{v}^{k+1} - map_{v}^{k}|
\end{equation} 

where $v \in \{f,s,t\}$ denotes the three projected views in the corresponding orthogonal Cartesian planes and $map_{v}^{k}$ is the projection of the k$th$ frame under the projection view $v$. In this way, temporal information which denotes the human body posture at the beginning of the action is captured together with the accumulated motion information, as shown in Fig~\ref{fig:DMM}(a)(b). In order to better exploit the motion patterns, DMMs are pseudo-colored into color images in the same way as that in~\cite{pichao2015}. In particular, the pseudo-color coding is done through the following hand-designed rainbow transform:

\begin{equation}\label{coloring}  
  C_{i={1,2,3}} = \{sin[2\pi\cdot(-I + \varphi_{i})\cdot\frac{1}{2}+\frac{1}{2}]\}^{2}\cdot f(I) 
\end{equation}
where $C_{i={1,2,3}}$ presents the BGR channels, respectively; $I$ is 
the normalized gray value; $\varphi_{i}$ denotes the phase of the three 
channels; $f(I)$ is an amplitude modulation function which further increases the non-linearity; the added values, $\frac{1}{2}$ guarantee non-negativity. We use the same parameter settings as~\cite{pichao2015}.

From Fig.~\ref{fig:DMM} it can seen that the two simple pair actions are almost the same in the traditional DMMs, but more discriminative in the modified DMMs. The temporal information in motion maps are enhanced through the pseudo-coloring method.

{\it ConvNets Training and Classification} Three ConvNets are trained on the pseudo-color coded DMMs constructed from the video segments in the three Cartesian planes. The layer configuration of the three ConvNets is same as the one in~\cite{krizhevsky2012imagenet}. The implementation is derived from the publicly available Caffe toolbox \cite{jia2014caffe} based on one {NVIDIA GeForce GTX TITAN X} card and the pre-trained models over ImageNet~\cite{krizhevsky2012imagenet} are used for initialization in training. 

For an testing action sample, only the original skeleton sequences without rotation are used to extract key frames for segmentation. Three DMMs are constructed from each segment in the three Cartesian planes as input to the ConvNets and the averages of the outputs from the three ConvNets are computed to label the testing video segments. The sequence of labels will serve as input to the subsequent HMM-SVM classifier. 

\subsection{HMM-SVM for Classification}
To effectively exploit the global temporal information, discrete HMMs are trained using the well-known Baum-Welch algorithm from the label sequences obtained from the ConvNets, one HMM per action. Specifically, the number of states are set to $K$, the number of clusters estimated from the IGMM, each state can emit one of the $K$ symbols. The observation symbol probability distribution is set as:

$$ B = \{b_{j}(k)\},\left\{
\begin{aligned}
b_{j}(k) = 1, &~ k = j \\
b_{j}(k) = 0, &~ k \neq j\\
\end{aligned}
\right.
$$
For each action sample, its likelihood being generated from the HMMs forms a feature vector as the input to the SVM for refining the classification.



{\small
\bibliographystyle{ieee}
\bibliography{myThesisBib}
}

\end{document}